\documentclass[10pt,twocolumn,letterpaper]{article}

\usepackage{cvpr}              

\usepackage[dvipsnames]{xcolor}

\definecolor{cvprblack}{rgb}{0.21,0.49,0.74}
\usepackage[pagebackref,breaklinks,colorlinks,citecolor=cvprblack]{hyperref}
\usepackage{bbding}
\usepackage{xcolor}
\usepackage{algorithm}
\usepackage{algpseudocode}

\newtheorem{theorem}{Theorem}

\newtheorem{definition}{Definition}
\newtheorem{property}{Property}

\usepackage{amssymb}
\usepackage{multirow}
\usepackage{booktabs}
\def\paperID{3595} 
\def\confName{CVPR}
\def\confYear{2024}

\title{ Data Valuation and Detections in Federated Learning}

\author{
  Wenqian Li \\
  National University of Singapore \\
  \texttt{wenqian@u.nus.edu} \\
   \and
  Shuran Fu \\
  NUS (Chongqing) Research Institution \\
  \texttt{shuran.fu@nusricq.cn} \\
    \and
  Fengrui, Zhang \\
  Georgetown University\\
  \texttt{fz128@georgetown.edu} \\
  \and
  Yan Pang\thanks{Corresponding Author} \\
  National University of Singapore \\
  \texttt{jamespang@nus.edu.sg} \\
}

\begin{document}
\maketitle
\begin{abstract}
Federated Learning (FL) enables collaborative model training while preserving the privacy of raw data. A challenge in this framework is the fair and efficient valuation of data, which is crucial for incentivizing clients to contribute high-quality data in the FL task. In scenarios involving numerous data clients within FL, it is often the case that only a subset of clients and datasets are pertinent to a specific learning task, while others might have either a negative or negligible impact on the model training process. This paper introduces a novel privacy-preserving method for evaluating client contributions and selecting relevant datasets without a pre-specified training algorithm in an FL task. Our proposed approach, \texttt{FedBary}, utilizes Wasserstein distance within the federated context, offering a new solution for data valuation in the FL framework. This method ensures transparent data valuation and efficient computation of the Wasserstein barycenter and reduces the dependence on validation datasets. Through extensive empirical experiments and theoretical analyses, we demonstrate the advantages of this data valuation method as a promising avenue for FL research. Codes are available at \href{https://github.com/muz1lee/MOTdata}{https://github.com/muz1lee/MOTdata}.
\end{abstract}   

\section{Introduction}
Federated Learning (FL) has emerged as a privacy-preserving approach for collaboratively training models without sharingw data~\cite{mcmahan2017communication}, in which the learning proceeds by iteratively exchanging model parameters between the server and clients. The success of the trained model hinges on the availability of large, high-quality, and relevant data~\cite{just2023lava}. Thus, it is crucial for the server to select valuable clients in the FL training process to ensure better model performance and explainability~\cite{sim2022data}. In the context of cross-silo FL, there is a mutual interest among clients in understanding the value of their own data as well as that of others. If there are free riders or malicious actors in FL, clients are unwilling to provide high-quality data to train the local model. Thus, fair valuation of data quality and accurate detection of noisy data become indispensable components of federated training~\cite{liu2022gtg}. 

In the context of a data marketplace with numerous potential clients, each possessing unique datasets, a central server issuing a learning task must discern and select the most valuable data clients for participation. This necessitates a fair and accurate assessment of each client's contribution. Commonly, existing methods evaluate contributions based on validation performance of the trained model, utilizing the concept of Shapley value (SV) to measure each client's marginal contribution to an FL model. This method, however, involves evaluating all possible subset combinations, leading to a computational complexity of $\mathcal{O}(2^N)$ for $N$ clients. Although efforts have been made to reduce the complexity through approximations~\cite{ghorbani2019data,jia2019towards,song2019profit,liu2022gtg}, these approaches may introduce biased evaluations and remain impractical for large-scale settings. Moreover, approaches based on validation performance are post-hoc, which assess client contributions after model training. This approach is problematic because some clients might be irrelevant to the FL task, and subsets of noisy data might also be involved in training without detection, leading to the waste of computational resources and poor model performance. Additionally, this evaluation method is typically client-level, based on model gradients, and does not delve into the granularity of individual data points, leading to a lack of transparency in the evaluation process.

Consequently, alternative strategies have been explored for evaluating clients and their data prior to model training. For instance, \cite{tuor2021overcoming} proposed a pre-training approach where small sizes of data samples are shared to achieve this. Recently, Wasserstein distance~\cite{alvarez2020geometric,just2023lava,kang2023performance} have been introduced to evaluate data without specifying a learning algorithm in advance. However, these approaches require access to data samples, which may not be feasible in privacy-sensitive settings. Moreover, they also rely on the validation dataset for assessment. Given challenges inherent in previous works, our research is driven by following questions:
\\
\textbf{\textit{1)~How to evaluate and select valuable data points without sharing any raw data?}} Previous methods can reasonably assess clients' contributions, there is a lack of a well-established approach to achieving a comprehensive understanding of individual data point contributions in FL. This limitation hinders transparency and the persuasive power of evaluations. Furthermore, developing such an approach would help the server select the most valuable data for model training.
\\
\textbf{\textit{2)~How to predict and evaluate data client contributions without involving model training?}} 
Previous approaches involve training federated models to evaluate validation performance or utilize generative models to learn data distributions~\cite{sim2020collaborative}. These methods are often computationally intensive and reliant on validation data. Addressing this, the key challenge is to develop data valuation methods that do not require direct training of federated models. Such an advancement is not merely beneficial but crucial, particularly in FL scenarios characterized by complex model training processes or where validation datasets are not readily available on the FL server.
\\
\textbf{\textit{3)~How to evaluate data contributors in a large-scale setting?} } As the number of clients increases, the complexity of previous methods based on Shapley value (SV) grows exponentially. Therefore, offering a lower-complexity data valuation method that does not require full data access will be fundamental for large-scale FL settings, especially those involving more than 100 clients.

Leveraging the advances in computational optimal transport~\cite{peyre2019computational} and recent developments in the federated scenario as in \cite{rakotomamonjy2023federated}, we introduce \texttt{FedBary} as an innovative solution to address the aforementioned challenges. To the best of our knowledge, FedBary is the first established general privacy-preserving framework to evaluate both client contribution and individual data value within FL. We also provide an efficient algorithm for computing the Wasserstein barycenter, to lift dependence on validation data. FedBary offers a more transparent viewpoint for data evaluations in FL, which directly computes the distance among various data distributions, mitigating the fluctuations introduced by model training and testing. Furthermore, it demonstrates faster computations without sacrificing performance compared to traditional approaches. We conduct extensive experiments and theoretical analysis to show the promising applications of this research. A comparison of representative state-of-the-art approaches is shown in Table~\ref{overview}.
\begin{table}
\small
\renewcommand{\arraystretch}{1}
\begin{tabular}{|c|c|c|c|c|}
\hline
\multirow{2}{*}{\textbf{Methods}}           & \multirow{2}{*}{Privacy}   & Without & Without & Noisy  \\&    &      validation          &  model     &  detection\\ \hline
DataSV~\cite{ghorbani2019data} &   $\checkmark$              &       \scalebox{0.85}[1]{$\times$}      &    \scalebox{0.85}[1]{$\times$}       &    \scalebox{0.85}[1]{$\times$}     \\   \hline
CGSV~\cite{xu2021gradient}  &  $\checkmark$               &       $\checkmark$      &     \scalebox{0.85}[1]{$\times$}      &     \scalebox{0.85}[1]{$\times$}    \\ \hline
GTG-SV~\cite{liu2022gtg}  &      $\checkmark$           &      \scalebox{0.85}[1]{$\times$}      &        \scalebox{0.85}[1]{$\times$}    &     \scalebox{0.85}[1]{$\times$}     \\ \hline
Lava~\cite{just2023lava}        &          \scalebox{0.85}[1]{$\times$}       &      \scalebox{0.85}[1]{$\times$}       &    $\checkmark$      & $\checkmark$       \\ \hline
Ours (FedBary)&   $\checkmark$             &    $\checkmark$        & $\checkmark$         &   $\checkmark$     \\ \hline
\end{tabular}
\vspace{-1.1em}
\caption{Overview of different approaches: we aim to handle both client evaluation and data detection tasks, w/o validation data}
\label{overview}
\vspace{-2.5em}
\end{table}
\section{Related Work}


\textbf{Optimal Transport Application} Optimal Transport (OT) is a classical mathematical framework used to solve transportation and distribution problems \cite{villani2009optimal,villani2021topics}, and it has been used in the field of machine learning for different tasks, including model aggregation and domain adaption \cite{flamary2016optimal}. Its effectiveness has been proved empirically \cite{courty2017joint} and theoretically \cite{redko2017theoretical,hoffman2018algorithms}. \cite{alvarez2020geometric} showed that it can be used to measure the distance between two datasets, providing a meaningful comparison of datasets and correlating well with transfer learning hardness . In order to get closer to a real situation, some research put their emphasis on multi-source domain adaption (MSDA), where there are multiple source domains and a robust model is required to perform well on any target mixture distribution \cite{hoffman2018algorithms}. 
The results of recent research showed that OT theory is capable of being used as a tool to solve multi-source domain adaption problems \cite{redko2019optimal,rakotomamonjy2022optimal,turrisi2022multi,wang2023class,montesuma2021wasserstein}. \\
\textbf{Data Valuation} The topic of data quality valuation has become popular and gained research interest in recent years since the quality of data will have a direct impact on the trained models, thus influencing downstream tasks. The most popular underlying metric for data valuation algorithms is the Shapley value \cite{jia2019towards,ghorbani2019data,kwon2021efficient}, which calculates the marginal contribution and measures the average change in the predefined function when a particular data point is removed. This function is commonly set as the performance of a model trained on a subset of the training dataset. Other papers following this method include KNNShapley \cite{jia2019efficient}, Volume-based Shapely \cite{xu2021validation}, BetaShapley \cite{kwon2021beta}, DataBanzhaf \cite{wang2023data} and AME \cite{lin2022measuring}. Alternative underlying methods are also explored to evaluate data contribution. For example, gradient-based approaches like Influence Function \cite{feldman2020neural} are proposed to quantify the rate of changes for a utility value when some data points are more weighted. \cite{pruthi2020estimating} proposed \emph{TracIn} method to estimate training data influence by tracing gradient descent. However, this method relies heavily on the training algorithms of machine learning models. \cite{just2023lava} developed a proxy for the validation performance, and their method Lava can be used to evaluate data in a way that is oblivious to the learning algorithms. Reinforcement learning \cite{yoon2020data} and out-of-bag estimation \cite{kwon2023data} are also introduced as tools to evaluate data values. A benchmark for data valuation is provided in \cite{jiang2023opendataval}, which can be conveniently accessed by data users and publishers.

\section{Technical Preliminary}
\subsection{Wasserstein Distance and Barycenter}
We denote by $\mathcal{P}(X)$ the set of probability measures in $X$ and $\mathcal{P}_p(X)$ the subset of measures in $\mathcal{P}(X)$ with finite $p$-moment $p\geq 1$. For $P\in\mathcal{P}_p(X)$ and $Q\in \mathcal{P}_p(Y)$, with distance function $d(x,y)$, the $p$-Wasserstein distance $\mathcal{W}_p(P,Q)$ between the measures $P$ and $Q$ is defined as
\begin{equation}
\label{primal}
    \mathcal{W}_p(P,Q) = \Big(\inf_{\pi\in\Pi(P,Q)}  \int_{X\times Y}d^{p}(x,y) d\pi(x,y) \Big)^{1/p}
\end{equation}
here $\Pi(P,Q)$ denotes the set of all joint distributions on $(X,Y)$ that are marginally distributed as $P$ and $Q$, and $\mathcal{W}_p(P,Q)$ quantifies the optimal expected cost of mapping samples from $P$ to $Q$. When the infimum in~\eqref{primal} is attained, any probability $\pi$ that realizes the minimum is an optimal transport plan. 
The notion of Wasserstein barycenter can be viewed as the mean of probability distributions $P_i, i\in[1, N]$ in the Wasserstein space~\cite{agueh2011barycenters}, defined as 
\begin{definition}
\label{defbarycenter}
    Given $P_i$ with $P_i\in \mathcal{P}_p(X), \forall i\in[1,N]$, with positive constants $\{\lambda_i\}_{i=1}^N$ such that $\sum_i^N\lambda_i =1$, the Wasserstein barycenter is the optimal $Q^\star$ satisfying
\begin{equation}
    Q^\star = \arg\min \sum_{i=1}^N \lambda_i\mathcal{W}_p(P_i,Q).
\end{equation}
\end{definition}
The problem of calculating the Wasserstein barycenter
of empirical measures $\hat{P}_i$ is proposed by~\cite{cuturi2014fast} since we can only access samples following distributions of corresponding $P_i$ in common case. 
In addition, consider the infimal convolution cost in the multi-source optimal transport theorey~\cite{pass2015multi}, $p$-Wasserstein distance with an $N$-ary distance function $d(x_1,\cdots,x_N)$ among all distributions $P_i$ is equivalent to find the minimal of the total pairwise between $Q$ and $P_i$, that is  $ \mathcal{W}_p(P_1,\cdots, P_n) = \min_Q \sum_{i=1}^N \mathcal{W}_p(Q, P_i)$, which provides a viewpoint of the mixture distribution of different source distributions and help measure the heterogeneity among them.

\subsection{ Geodesics and Interpolating Measure}
This part hinges on the geometry of the Wasserstein distance and geodesics for applying in a federated manner~\cite{rakotomamonjy2023federated}. We utilized the triangle inequality property of Wasserstein distance and defined the concept of geodesics and the interpolating measure. We show the details of the definitions in Appendix ~\ref{geodesic_def}.

Denote $\bar{\mu}$ as the geodesics, then any point $\gamma_t$ on $\bar{\mu}$ is an interpolating measure between distribution $P$ and $Q$, thus formulating the equality:
\begin{equation}
\label{geodesicseq}
    \mathcal{W}_p(P,Q)  = \mathcal{W}_p(P,\gamma_t) + \mathcal{W}_p(\gamma_t,Q).
\end{equation}
This provides insight for computing the Wasserstein distance in a federated manner: once we can approximate the interpolating measure between two distributions, we can measure the Wasserstein distance based on~\eqref{geodesicseq}.

\begin{figure}
\setlength{\abovecaptionskip}{0.1 cm} 
\setlength{\belowcaptionskip}{-0.6cm} 
\vspace{-0.5cm} 
\centering
    \includegraphics[width =0.45\textwidth]{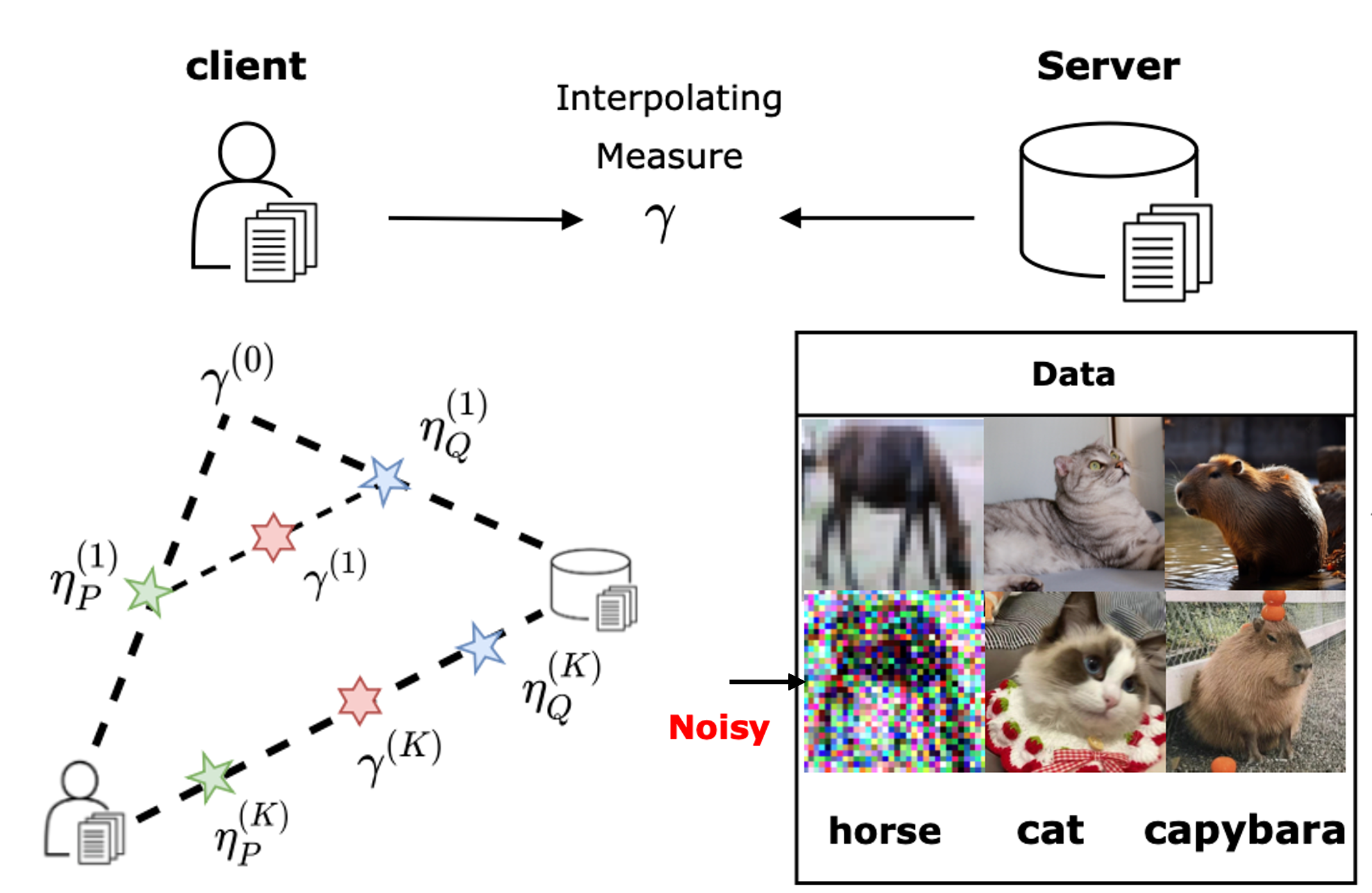}
    \caption{Client holds $P$ and server holds $Q$, the interpolating measure $\gamma$ aids to measure distance $\mathcal{W}_p(P,Q)$. Local interpolating measures $\eta_p$ and $\eta_Q$ are shared for calculation and detection.}
\label{fedbary_fig}
\end{figure}
\section{Problem Formulation}
Suppose there are $N$ benign clients and each client $i$ holds dataset $\mathcal{D}_i = \{(x_{i,j},y_{i,j})\}_{j=1}^{m}$ with size $m$. Considering data heterogeneity, we assume $\mathcal{D}_i$ is independently and identically sampled from the different distribution $P_i$, but shares the same feature space $\mathcal{X}$ and label space $\mathcal{Y}$ such that $\mathcal{X} \times \mathcal{Y}=\mathcal{Z}$. Since we could not access the true distribution, in general, one can construct discrete measures $P_i(x,y):= \frac{1}{m}\sum_{j=1}^m \delta_{(x_{i,j},y_{i,j})}$, where $\delta$ is a Dirac function. Clients are supposed to collaboratively train a model, and before training, the server wants to measure the contribution and select relevant data to a target distribution $Q$. If the server holds validation dataset $\mathcal{D}_Q =  \{(x_{q,j},y_{q,j})\}_{j=1}^{m}$, we assume is i.i.d to $Q$. If the server does not have a validation dataset, the target distribution is approximated by $\hat{Q}$. We use the Euclidean distance $d(\cdot,\cdot)$ to measure feature distance. For the label distance, we use the conditional feature space as $P_i(x|y_o) = \frac{P_i(x)\mathbb{I}[y=y_o]}{\int P_i(x)\mathbb{I}[y=y_o]dx}$ with $y_o \in \mathcal{Y}$~\cite{alvarez2020geometric}. In order to calculate the Wasserstein distance $\mathcal{W}_p(P_i,Q)$, for each client, there is an interpolating measure $\gamma_i$ to be approximated. Therefore, $\gamma_i$ will be initialized and shared between the server and client $i$ for iteratively updating. Any raw dataset $D_Q$ and $D_i,\forall i\in[1,N]$ will not be shared. 

In the following sections, we will first provide the technique to approximate Wasserstein barycenter in Section~\ref{fedbary} for a better understanding of the evaluating procedure. Then, we will investigate the application of Wasserstein measure in scenarios with and without validation datasets in Section~\ref{dataval}. Furthermore, we leverage the duality theorem to detect noisy and irrelevant data points in Section~\ref{detectdata}. Figure~\ref{fedbary_fig} shows the overall framework.

\subsection{Federated Wasserstein Barycenter}
\label{fedbary}
Our goal is to approximate the Wasserstein barycenter in Definition~\ref{defbarycenter} among data distributions $P_1, P_2, \cdots, P_N$ on the server. Without loss of generality, we assume $\lambda_i =1 $.
In order to make the update $\gamma_i$, the triangle inequality defined in Property~\ref{triangle inequality} is extended in the following way,
\begin{align}
\label{foursum}
    &\mathcal{W}_p(P_i,Q) \leq \mathcal{W}_p(P_i,\gamma_i)+\mathcal{W}_p(\gamma_i,Q), \nonumber\\
    &\mathcal{W}_p(P_i,\gamma_i)= \mathcal{W}_p(P_i,\eta_{P_i})+\mathcal{W}_p(\eta_{P_i},\gamma_i),\nonumber\\
    &\mathcal{W}_p(\gamma_i,Q)= \mathcal{W}_p(\gamma_i,\eta_{Q_i})+\mathcal{W}_p(\eta_{Q_i},Q),
\end{align}
where $\eta_{P_i}$ is the interpolating measure between $P_i$ and $\gamma_i$ computed by $i$-th client, and $\eta_{Q_i}$  is the interpolating measure between $\gamma_i$ and $Q$ computed by the server. The interpolating measure between $P_i$ and $\gamma_i$ (same way to $Q$ and $\gamma_i$) could be approximated~\cite{rakotomamonjy2023federated} based on
\begin{equation}
\label{inter_measure}
    \eta_{P_i} = \frac{1}{m}\sum_{j=1}^m\delta_{(1-t)x_j+tm(\pi^\star\mathbf{Z}^\prime)_j},
\end{equation}
where $\pi^\star$ is the optimal transportation plan between $P_i$ and $\gamma_i$, $x_j$ is $j$-th sample from $P_i$, and $\mathbf{Z}^\prime$ is the matrix of samples from $\gamma_i$. $t\in[0,1]$ is a hyperparameter.

Once $\gamma_i$ is an interpolating measure between $P_i$ and $Q$, $\mathcal{W}_p(P_i,Q)$ equals the summation of four terms on the right-hand side in~\eqref{foursum} and could be further applied to approximate the barycenter when $\min \sum_{i=1}^N\mathcal{W}_p(P_i,Q)$ is attained. Therefore, we develop a $K$-round iterative optimization procedure to approximate the interpolating measure $\gamma_i^{(k)}$ and the Wasserstein barycenter $Q^{(k)}$ based on $\eta_{P_i}^{(k)}$ and $\eta_{Q_{i}}^{(k)}$. Specifically, at each iteration $k$, the clients receive current iterate $\gamma_i^{(k-1)}$ and compute interpolating measure $\eta_{P_i}^{(k)}$ between $P_i$ and $\gamma_i^{(k-1)}$, and server computes interpolating measure $\eta_{Q_{i}}^{(k)}$ between each $\gamma_i^{(k-1)}$ and $Q^{(k-1)}$. 
Then the server sends all $\eta_{Q_{i}}^{(k)}$ to clients, and $i$-th client computes the next iterate $\gamma_i^{(k)}$ based on $\eta_{P_i}^{(k)}$ and $\eta_{Q_i}^{(k)}$. The distance in an iterative procedure is calculated based on~\eqref{iter_cal} as follows,
\begin{align}
\label{iter_cal}
    \mathcal{W}_p(P_i,Q) \leq \mathcal{W}_p(P_i,\eta_{P_i}^{(k)})+\mathcal{W}_p(\eta_{P_i}^{(k)},\gamma_i^{(k-1)})\nonumber\\+\mathcal{W}_p(\gamma_i^{(k-1)},\eta_{Q_{i}}^{(k)})+\mathcal{W}_p(\eta_{Q_{i}}^{(k)},Q^{(k-1)}),
\end{align}
\textcolor{black}{where $\gamma_i^{(k)}$ is updated as~\cite{rakotomamonjy2023federated} by}
\begin{equation}
\label{gammaupdate}
\gamma_i^{(k)} \in \texttt{argmin}~ \big[ \mathcal{W}_p(\eta_{P_i}^{(k)},\gamma_i^{(k-1)})+\mathcal{W}_p(\gamma_i^{(k-1)},\eta_{Q_i}^{(k)})\big]. 
\end{equation}
It is straightforward to see the distribution $Q$ is only involved in $\mathcal{W}_p(Q^{(k-1)}, \eta_{Q_i}^{(k)})$ 
and thus we can easily update barycenter $Q^{(k)}$ simultaneously by transporting samples from $\gamma_i^{(k)}$ to the common distribution.

As for the pairwise $z=(x,y)$, it is not easy to approximate the interpolating measure with classification labels. \textcolor{black}{Based on the insight of \cite{alvarez2020geometric,rakotomamonjy2023federated}} , we utilize the point-wise notion of distance in $\mathcal{X}\times \mathcal{Y}$ as 
\begin{align}
\label{pointdistance}
    &d\big((x,y),(x^\prime,y^\prime)\big)\triangleq \big(d(x,x^\prime) + \mathcal{W}_2^2 (\alpha_y,\alpha_{y^\prime}) \big)^{1/2},\nonumber\\
    &\mathcal{W}_2^2 (\alpha_y,\alpha_{y^\prime}) = ||m_y-m_{y^\prime}||_2^2 + ||\Sigma_y-\Sigma_{y^\prime}||_2^2,
\end{align}
where $\alpha_{y^\prime}$ is conditional feature distribution $P(x|y=y^\prime)$ that follows the Gaussian distribution with mean $m_y$ and covariance $\Sigma_y$, we can construct the augmentated representation of each dataset, that each pair $(x,y)$ is a stacked vector $\tilde{x}:=[x;m_y;\text{vec}(\Sigma_y^{1/2})]$. Then with the stacked matrix $\mathbf{\tilde{X}}$ and $\mathbf{\tilde{X}^\prime}$, the data distance is calculated based on $\mathcal{W}_p(\mathbf{\tilde{X}},\mathbf{\tilde{X}^\prime})\leq \mathcal{W}_p(\mathbf{\tilde{X}},\gamma)+ \mathcal{W}_p(\gamma,\mathbf{\tilde{X}^\prime}) $. We use $\mathbf{\tilde{X}}_i$ and $\mathbf{\tilde{X}}_Q$ as the stacked vector for the samples from $P_i$ and $Q$.
Our algorithm is summarized in Algorithm~\ref{algorithm_1_fedbary}.

\subsection{Evaluate Client Contribution}
\label{dataval}
When the server has a validation set $D_Q =\{x_{q,j},y_{q,j}\}_{j=1}^m$, then it can easily measure $\mathcal{W}_p(\mathbf{\tilde{X}}_i,\mathbf{\tilde{X}}_Q)$  without the initialization of $Q^{(0)}$ and update on $\mathbf{\tilde{X}}_Q^{(k)}$(line 7 in Algorithm~\ref{algorithm_1_fedbary}). 
From \cite{just2023lava}, we know the validation loss of the trained model is bounded by the distance between the training data and the validation data. Thus, we can measure the contribution of $i$-th using the reverse of Wasserstein distance $\mathcal{W}_p(\mathbf{\tilde{X}}_i,\mathbf{\tilde{X}}_Q)$ without training a federated model. A smaller distance leads to better performance on the validation data $D_Q$ and can be considered more valuable.

\begin{algorithm}[h]
    \caption{FedBary} %
  \label{algorithm_1_fedbary}
  \begin{algorithmic}[1]
  \renewcommand{\algorithmicrequire}{\textbf{Input:}}
\renewcommand{\algorithmicensure}{\textbf{Output:}}
    \Require
      Local data distribution $P_i$ with $\mathbf{\tilde{X}}_i$, initialisation of $Q^{(0)}$ with support $\mathbf{\tilde{X}}_Q^{(0)}$ and  $\gamma_i^{(0)}, i=1,\cdots N.$
      \textcolor{black}{(No initialisation of $\mathbf{\tilde{X}}_Q^{(0)}$ with validation set, use fixed $\mathbf{\tilde{X}}_Q$ )}
    \For{$k=1$ to $K$}
        \State Clients compute distance $\mathcal{W}_p(\mathbf{\tilde{X}}_i, \gamma_i^{(k-1)})= \mathcal{W}_p(\mathbf{\tilde{X}}_i,\eta_{P_i}^{(k)})+\mathcal{W}_p(\eta_{P_i}^{(k)},\gamma_i^{(k-1)})$
        \State Clients send $\mathcal{W}_p(\mathbf{\tilde{X}}_i, \gamma_i^{(k-1)})$ and $\gamma_i^{(k-1)}$ to server.
        \State Server computes distance $\mathcal{W}_p(\mathbf{\tilde{X}}_Q^{(k-1)}, \gamma_i^{(k-1)})= \mathcal{W}_p(\mathbf{\tilde{X}}_Q^{(k-1)},\eta_{Q_i}^{(k)})+\mathcal{W}_p(\eta_{Q_i}^{(k)},\gamma_i^{(k-1)})$
        \State Server sends $\eta_{Q_i}^{(k)}$ to $i$-th client
        \State Client $i$ updates $\gamma_i^{(k)}$ based on $\eta_{P_i}^{(k)}$ and $\eta_{Q_i}^{(k)}$.
        \State Server updates $\mathbf{\tilde{X}}_Q^{(k)}$ \textcolor{black}{ (If without validation set) }
    \EndFor 
    \Ensure $\mathbf{\tilde{X}}_Q^{(K)}, \gamma_i^{(K)}, \eta_{P_i}^{(K)},\eta_{Q}^{(K)},\mathcal{W}_p(\mathbf{\tilde{X}}_Q^{(K)}, \mathbf{\tilde{X}}_i)$
  \end{algorithmic}
\end{algorithm}
 However, if there is no validation dataset $D_Q$, the intialisation of $Q^{(0)}$ with $\mathbf{\tilde{X}}_Q^{(0)}$ should follow the same dimension of constructed matrix $\mathbf{\tilde{X}}_i$ using $D_i$. Some work \cite{mohri2019agnostic,reisizadeh2020robust} discuss that the target distribution~(the model truly learns)~is the mixture of local distributions, i.e., $\mathcal{U}_\mathbf{\lambda} = \sum_{i=1}^N \lambda_i P_i$ for some $\mathbf{\lambda} \in \triangle_N, \triangle_N:=\{\mathbf{p}\in [0,+\infty)^N:\langle~\mathbf{p},\mathbf{1}_N ~\rangle =1\}$, which is equivalent to the $\lambda$-weighted Euclidean barycenter of the distribution $P_1,\cdots,P_N$. \textcolor{black}{In this paper, instead of the Euclidean barycenter, we opt for the Wasserstein barycenter $\lambda\mathcal{W}p(\mathcal{U},P_i)$ as the target distribution to evaluate the distance for two main reasons. Firstly, the Wasserstein barycenter is superior in capturing complex geometric structures. Secondly, employing a similar approach to approximate the Federated Euclidean barycenter poses a risk of data exposure. This risk arises from the inherent simplicity of the weighted averaging process.} Due to some fluctuations and the different sampling schemes in the various FL algorithms, coefficients $\lambda$ may vary and cause a mismatch between the assumed distribution and the true distribution. Thereby it is beneficial to consider any possible $\lambda$ for robust evaluations. We encourage more exploration of the choice of $\lambda$ for the future work.
 \subsection{Datum Detection}
 \label{detectdata}
\begin{figure}
\setlength{\abovecaptionskip}{0.1cm}
\setlength{\belowcaptionskip}{-0.6cm}
\vspace{-0.5cm}
\centering
\includegraphics[width=0.45\textwidth,height=0.23\textwidth]{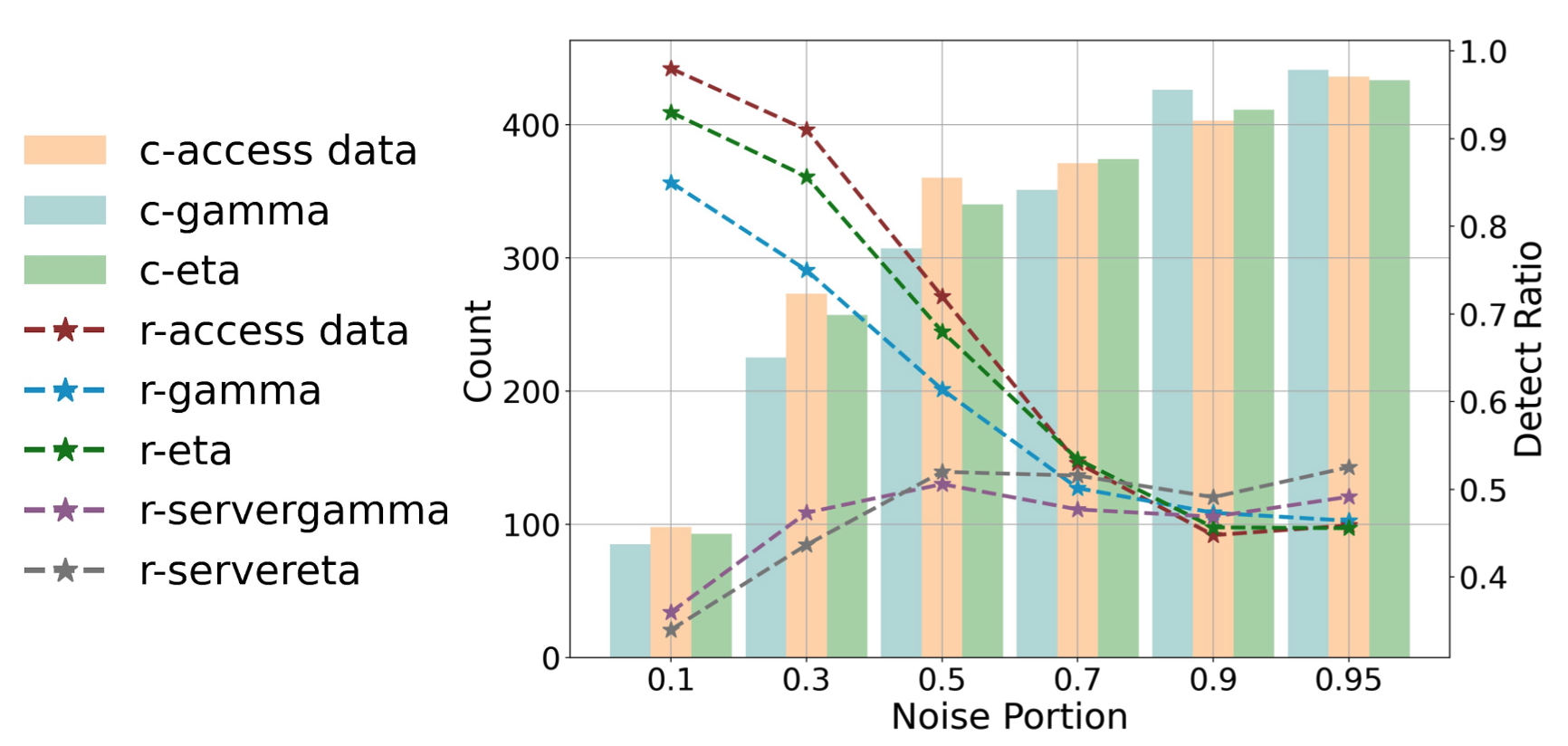}
\caption{We compare 5 different approaches for shuffled data detections: \textbf{gamma} corresponds to $\partial \mathcal{W}_p(P_i,\gamma_i)$; \textbf{eta} corresponds to $\partial \mathcal{W}_p(P_i,\eta_{Q_i})$; \textbf{access data} corresponds to $\partial \mathcal{W}_p(P_i,Q)$ (access both datasets). \textit{Noise Portion} represents the actual noisy data ratio in a dataset, \textit{count}~(c-)   indicates the number of detected negative calibrated gradient values;  \textit{Detection ratio}~(r-) measures detection accuracy: ($\#$detected noisy data $\slash$ $\#$noisy data); \textbf{servergamma}~(gray) and \textbf{servereta}~(purple) lines corresponds to $\partial \mathcal{W}_p(\eta_{P_i},\eta_{Q_i})$ and $\partial \mathcal{W}_p(\eta_{P_i},Q)$.}

\label{Attack Detections}
\end{figure}

Inspired from~\cite{just2023lava}, by leveraging the duality theorem, we could derive the dual problem $\mathcal{W}_p(P_i,Q):=\max_{(f,g)\in C^0(\mathcal{Z})^2}\langle~f, P_i~\rangle+\langle~g, Q~\rangle $ of the primal problem in~\eqref{primal}, where $C^0(\mathcal{Z})$ is the set of all continuous functions, $f,g\in \mathbb{R}^{m\times 1}$ are dual variables. Strong duality theorem says if $\pi^\star$ and $(f^\star,g^\star)$ are optimal variables of the corresponding primal and dual problem respectively, then we have $\mathcal{W}_p(\pi^\star) = \mathcal{W}_p(f^\star,g^\star)$. We applied this theorem into our scenario, when $\gamma_i$ is the interpolating measure between $P_i$ and $Q$, we have two separate dual problems for $\mathcal{W}_p(P_i,\gamma_i)$ and $\mathcal{W}_p(\gamma_i,Q)$, where 
\begin{align}
\label{dual}
    \mathcal{W}_p(P_i,Q) = \max_{(f,g)\in C^0(\mathcal{Z})^2}\langle~f,P_i~\rangle+\langle~g,\gamma_i~\rangle \nonumber \\
    +\max_{(h,j)\in C^0(\mathcal{Z})^2}\langle~h,\gamma_i~\rangle+\langle~j,Q~\rangle.
\end{align}
Therefore, we can get $ \partial_{P_i} \mathcal{W}_p(P_i,Q) \approx \partial_{P_i}\mathcal{W}_p(P_i,\gamma_i) = (f^\star)^T$, which is the gradient of the distance w.r.t. the distribution $P_i$. We could measure the quality of each datum using the \textit{calibrated gradient} as follows,
\begin{align}
\label{calivalue}
    \frac{\partial \mathcal{W}_p(P_i,\gamma_i)}{\partial P_i(z_l)} = f_l^\star - \sum_{{j\in \{1,\cdots,m\}\backslash l }}\frac{f_j^\star}{m-1},
\end{align}
which represents the rate of change in $\mathcal{W}_p(P_i,\gamma_i)$ w.r.t. the given datum $z_l$ in $D_i$. \\
\textbf{Interpretation}~ This value is interpreted as the contribution of a specific datum to the distance since it determines the shifting direction based on whether it is positive or negative. If the value is positive$\slash$negative, shifting more probability mass to that datum will result in an increase$\slash$decrease of the distance between the local distribution and the interpolating measure, further resulting in an increase$\slash$decrease of the distance between the local and the target distribution.\\
\textbf{Accurate detections} Since $\gamma_i$ is the approximated interpolating measure sampling from both $P_i$ and $Q$, then any function involved $\gamma_i$ in~\eqref{dual} also provides gradient information of $P_i$. However, with fixed $Q$, the interpolating measure will shift only if $P_i$ is changed, and in our empirical exploration in Figure~\ref{Attack Detections}, since the server shares $\eta_{Q_i}$ to the client,  $\partial\mathcal{W}_p(P_i,\eta_{Q_i})$ detects better than $\partial\mathcal{W}_p(P_i,\gamma_i)$.

\subsection{Theoretical Analysis}
In this section, we will provide the theoretical insights to justify our approach. \textcolor{black}{The convergence guarantee and complexity analysis are similar to~\cite{rakotomamonjy2023federated} based on the computation of Wasserstein distance in a Federated scenario.} \\
\textbf{Convergence Guarantee} First we will show that Algorithm \ref{algorithm_1_fedbary} has a convergence guarantee.
\begin{theorem}
\label{theorem1}
Let $P_i$ be the distribution of $i$-th client, where $i \in [1,N]$, and $Q^{(k)}$ be the Wasserstein barycenter at iteration $k$, $\gamma_i^{(k)}, \eta_{P_i}^{(k)}, \eta_{Q_i}^{(k)}$ be the interpolating measures computed in the Algorithm \ref{algorithm_1_fedbary}. Define
\begin{equation}
     A^{(k)} = \sum_i^N \Big[\mathcal{W}_p(P_i, \gamma_i^{(k)}) + \mathcal{W}_p(Q^{(k)}, \gamma_i^{(k)})\Big],
\end{equation}
then, the sequence $(A^{(k)})$ is non-increasing and converges to $\sum_{i=1}^N\mathcal{W}_p(P_i, Q)$.
\end{theorem}
We refer the proof to Appendix~\ref{proof_theorem1} and also conduct toy experiments to verify it later.\\
\textbf{Complexity Analysis} FedBary algorithm computes $3$ OT plans per iteration for evaluating one client, $2$ for the client and $1$ for the server. If there are $N$ clients, the server totally calculates $N\times K$ OT plans. Each OT plan's complexity is based on the network simplex algorithm, which operates at $\mathcal{O}( 2m^3log(2m))$ if balanced. 
Any interpolating measure between two distributions is supported by at most $2m+1$ points. Based on the approximation for the interpolating measure in~\eqref{inter_measure}, the overall computation cost is $\mathcal{O}(3NK(Sm^2+S^2)log(m+S))$ with the support size $S$ for $\gamma_i^{(k)}$. In Figure~\ref{wdfigure} in the Appendix, we could obverse different $S$ will not affect the contribution topology. Thus, we could reduce the complexity with small $S$. In real applications, FedBary might be appropriate, especially when $N$ is large since the complexity is linear with $N$. To show FedBary could be applied at scale, we compare the elapsed time for evaluating with different $N$ and $S$ in Table~\ref{compare_time}.\\
\textbf{Performance Bound} 
Without any downstream training data detections, the Wasserstein distance $\mathcal{W}_p(P_i,Q)$ can be simply used as the proxy for the validation performance when $D_Q$ is available. If $D_Q$ is not available on the server, we assume that $D_Q$ can be drawn i.i.d from the Wasserstein barycenter $Q$ approximated by Algorithm \ref{algorithm_1_fedbary}.
\begin{theorem}(\cite{just2023lava})
\label{theorem2}
Denote $f_t:\mathcal{X}\rightarrow\{0,1\}^V$, $f_v:\mathcal{X}\rightarrow\{0,1\}^V$ as the labeling functions for training and validation data. where $V$ is the number of different labels. Let $f:\mathcal{X}\rightarrow\{0,1\}^V$ be the model trained on training data. 
Let $P_i(\cdot|y_o)$ and $Q(\cdot|y_o)$ be the corresponding conditional distributions given label $y_o$. Assume that the model $f$ is $\epsilon$-Lipschitz and the loss function $\mathcal{L}:\{0,1\}^V \times [0,1]^V \rightarrow \mathbb{R}^+$ is $k$-Lipschitz in both inputs. Define distance function $d$ between $(x_i,y_i)$ and $(x_q,y_q)$ as~\eqref{pointdistance}. Under a certain cross-Lipschitzness assumption for $f_t$ and $f_v$, we have
    \begin{align}
     &\mathbb{E}_{x\sim Q(x)}[\mathcal{L}(f_v(x),f(x))]\\  &\leq
     \mathbb{E}_{x\sim P_i(x)}[\mathcal{L}(f_t(x),f(x))] 
     + k\epsilon\mathcal{W}_p(P_i,Q) + \mathcal{O}(\epsilon V)\nonumber 
    \end{align}
\end{theorem}
The proof is outlined in Appendix ~\ref{proof_theorem2}.
This theorem indicates validation loss is linearly changed with the  $\mathcal{W}_p(P_i,Q)$ if the training loss (first term on rhs) is small enough.\\
\textbf{Privacy Guarantee} \textcolor{black}{In our setting, privacy is guaranteed from two aspects: (1) Raw training samples cannot be reconstructed. (2) Generating data with similar appearance is not feasible. The reasons behind this are: Firstly, the stacked vector does not replicate the original data structure. Secondly, hyperparameter $t$ for approximating the interpolating measure, the OT plan between $\tilde{\mathbf{X}}_i$ and $\gamma_i$, and OT plan between $\eta_{Q_i}$ and $\eta_{P_i}$ are only known to the client $i$ himself. Therefore, it is challenging for attackers who need to know all the information to infer raw datasets.}


\section{Experiments}
\begin{figure}
\setlength{\abovecaptionskip}{0.1 cm} 
\setlength{\belowcaptionskip}{-0.5 cm} 
    \includegraphics[width =0.45\textwidth]{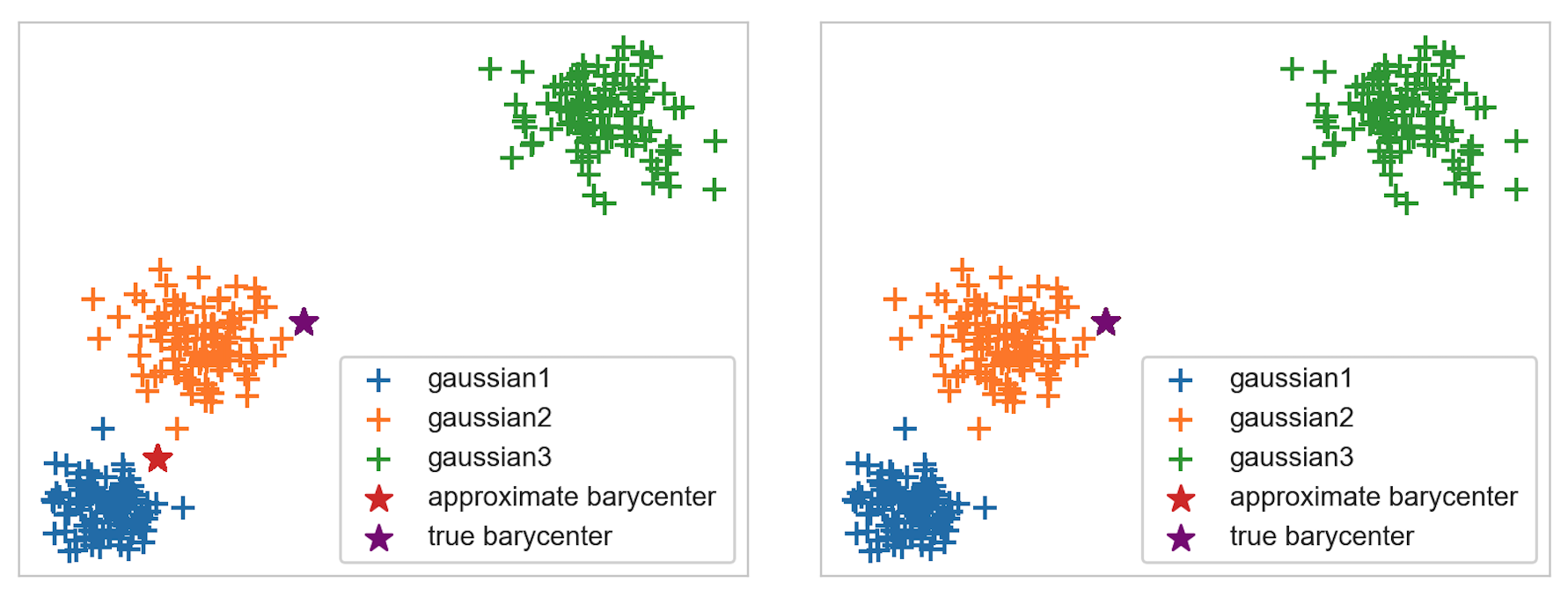}
    \caption{Approximated and true Wasserstein barycenter of 3 Gaussian distributions: 3-th epoch and 10-th epoch~(overlapping).}
\label{gaussian_toy}
\end{figure}

We demonstrate the computation of the Wasserstein barycenter among three Gaussian distributions from the empirical perspective to make Theorem~\ref{theorem1} more convincing. For each distribution, we sample 100 data points from 2D Gaussian distributions with distinct means but the same covariance matrix. We set $t=0.5$ for interpolating the measure, as approximated in Section~\ref{inter_measure}. To assess the accuracy of our computations, we compare the results obtained using FedBary with the barycenter approximation with data access as outlined in the work by~\cite{cuturi2014fast}. In our study, we quantified the disparity between the barycenter estimated within FL and the barycenter derived from accessible data by computing the squared errors for each position and subsequently aggregating all differences. Remarkably, our experimental results demonstrated a swift convergence, typically within a mere 10 iterations in Figure~\ref{gaussian_toy}. 

\begin{table}

    \centering
    \setlength{\tabcolsep}{0.8mm}
    \begin{tabular}{ccccccc}
    \toprule[1pt] 
     $N$ & \textbf{ExactFed} &\textbf{GTG} &\textbf{MR} & \textbf{DataSV} & \textbf{Ours}~($S=100\slash500$)  \\\hline
    $5$ &  31m & 33s	&  5m & 25m &1m $\slash$  2m  \\\hline   
    $10$ &  3h20m   &  7m	& 40m & 2h30m &2m  $\slash$  4m \\\hline   
    $50$ &  - &	-&  - & -&14m   $\slash$30m\\\hline
    $100$ & -  &- &- & -&21m  $\slash$ 1h  \\
    \bottomrule[1pt]     
    \end{tabular}
    \caption{Evaluation time with different size of $N$: For ExactFed, GTG and MR, we only consider the evaluation time after model training; Evaluation time of FedBary increases linearly with $N$. }
    \label{compare_time}
\end{table}
\subsection{Clients Evaluation}
\label{clienteva}
\textbf{Datasets} \textcolor{black}{We used the CIFAR-10 dataset in this section for experiments (more dataset experiments are shown in Appendix), and followed the data settings in~\cite{liu2022gtg}.} We simulate $N=5$ clients and randomly sample data for each client with 5 cases: \textit{(1)~Same Distribution and Same Size};\textit{(2)~Different Distributions and Same Size};\textit{(3)~Same Distribution and Different Sizes}; \textit{(4)~Noisy Labels and Same Size};\textit{(5)~Noisy Features and Same Size}. Refer Appendix~\ref{clientvaldata} for details.\\
\begin{figure}
\setlength{\abovecaptionskip}{0.1 cm} 
\setlength{\belowcaptionskip}{-0.5 cm} 
    \includegraphics[width =0.49\textwidth]{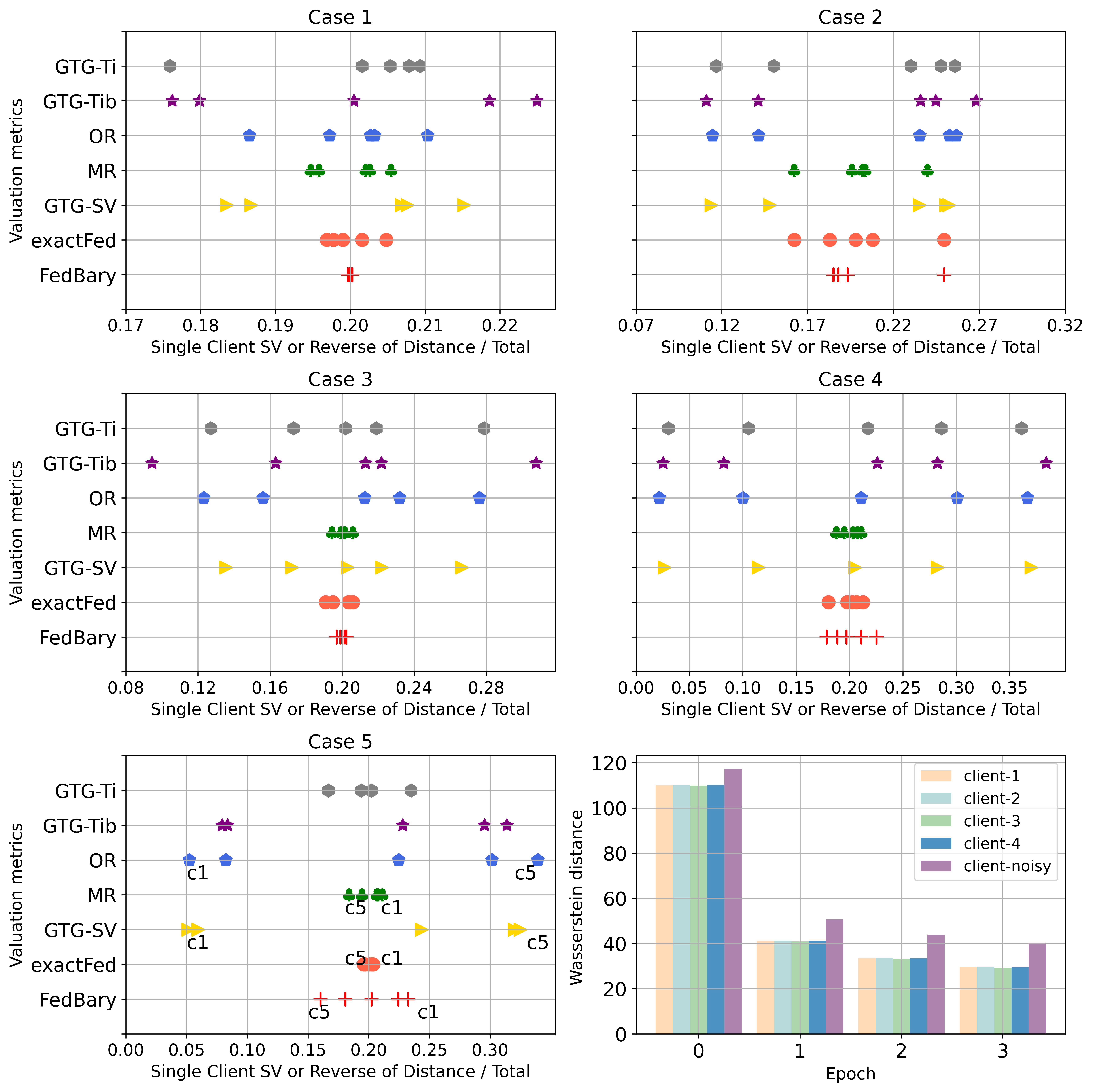}
    \caption{Scatter plots: percentage of contribution for 5 clients under different valuation metrics (Case1$\sim$5); Histogram: distance between the local distribution and the Wasserstein barycenter when validation set is not available.}
\label{case1_5}
\vspace{-0.2cm} 
\end{figure}
\textbf{With Validation data} When the validation data is available, our algorithm can be directly used to compute the distance between data from each client and the validation data, which is usually stored on the server side in reality. Thus, the shorter distance implies the better contribution from client data. We compared our method with other data valuation metrics. Baselines are \emph{GTG-Shapley} with its variants~(\emph{GTG-Ti$\slash$GTG-Tib})~\cite{liu2022gtg}, \emph{FedShapley} \cite{wang2020principled}, \emph{MR/OR}~\cite{song2019profit}, \emph{DataSV}~\cite{ghorbani2019data} and exact calculation \emph{exactFed}.

We visualize all cases in Figure~\ref{case1_5} to show the percentage of contribution when the number of clients equals 5 under different valuation metrics. The X-axis is the contribution score($\%$), and the y-axis are valuation approaches. Each marker stands for the score of a client. We measured the percentage of contribution by dividing each client's Shapley value or inverse of distance by the total. 
The exactFed provides the original Shapley value, which can be considered as ground truth to some extent.
For case 1, the result of our distance metric shows that the contributions for all 5 clients are almost equal, with each of them occupying around 20$\%$, which is a signal that this metric outperforms others since the datasets are i.i.d with the same size. For case 2, FedBary could differentiate distributions and follow similar contribution scores with exactFed and MR. For case 3, with the same distribution, the size of data samples will have a trivial influence on the contribution, which shows the robustness to replications of FedBary, while other approaches GTG-Ti,GTG-Tib, GTG-SV are sensitive to the data size. We can also find MR, which is shown to be more accurate than other SV approaches for evaluations,  achieves similar performance with ours and exactFed in this setting. If noise exists in labels of data~(case 4), we can expect that the higher the percentage of noisy labels, the smaller the contribution score is. Some metrics like GTG-SV will show clear discretization among clients with different percentages of noisy labels. Most of them range from 2\% to 38\%. However, the exactFed is not that dispersed, as the SVs among different clients are close. \textcolor{black}{FedBary follows this trend and delivers a result similar to exactFed and MR, showing its ability to differentiate the mislabeled data.} For case 5, FedBary outperforms other approximated approaches since they will have an inverse evaluation for the clients: with a larger proportion of noise, the contribution score is larger. FedBary is sensitive to feature noise and could capture the right ordering of client contributions: with a larger proportion of noise, the contribution score is smaller. Overall, FedBary provides better evaluations.\\
\textbf{Without Validation data} 
Wasserstein barycenter could assist in identifying irrelevant clients (distribution that is far from others) or distribution with noisy data points. Although there is another attempt by calculating $\mathcal{W}_p(P_i,P_j), i\neq j$ \cite{rakotomamonjy2023federated} to measure the data heterogeneity, such procedure needs to compute $\binom{N}{2}$ OT plans in each iteration, causing high computational cost. We simulate the scenario where there are $4$ clients with i.i.d distributions and the $5$-th client with noisy data, i.e., with noisy features in case~(5) above. Then we approximate the Wasserstein barycenter with support $\mathbf{\tilde{X}}_Q$ among these clients and measure the distance $\mathcal{W}_p(\mathbf{\tilde{X}}_i,\mathbf{\tilde{X}}_Q)$. We plot the result in Figure~\ref{case1_5}, in which we could find that the $\mathcal{W}_p(\mathbf{\tilde{X}}_5,\mathbf{\tilde{X}}_Q)$ is larger than other distances, leading to the conclusion that this dataset is relatively irrelevant.

\subsection{Noisy Feature$\slash$Mislabeled Data Detection}
\label{noisydata}
As aforementioned, our approach represents the first attempt in FL to perform noisy data detections without sharing any data samples. To gauge the accuracy and effectiveness of our algorithm, we conducted comparative evaluations with existing approaches that could access the dataset. Baselines here are \emph{Lava}, \emph{LOO} and \emph{KNNShapley}.

We follow the experiment setting in~\cite{jiang2023opendataval}, where we consider two types of synthetic noise: 1)~label noise where
we randomly assign the label that is different from the original label of the data points with a predefined proportion; 2)~feature noise where we add standard Gaussian random noises to the original features. Here we randomly choose the proportion $p_{\text{noise}}\%$ of the training dataset to perturb. Here we consider three different levels $p_{\text{noise}}\in\{5,15,30\}\%$. 
\begin{figure}
\setlength{\abovecaptionskip}{0.1 cm} 
\setlength{\belowcaptionskip}{-0.5 cm} 
\vspace{-0.5cm} 
\centering
    \includegraphics[width =0.5\textwidth]{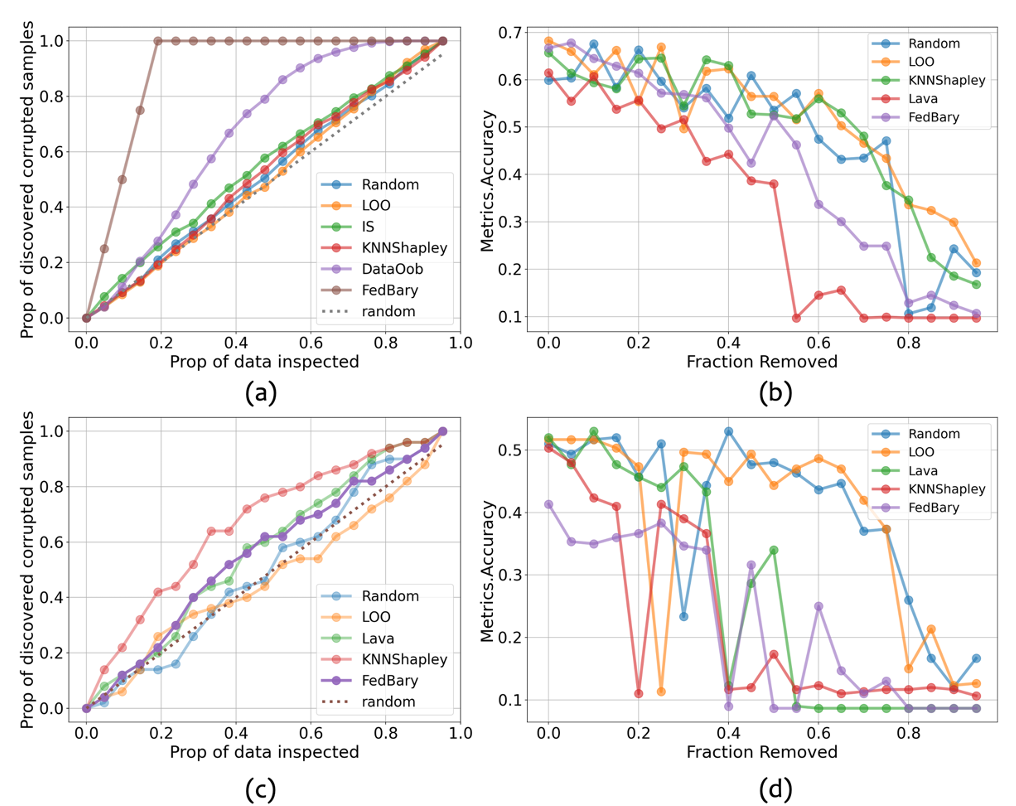}
        \caption{ \textcolor{black}{Detection results on CIFAR10: \textbf{Corrupt feature samples detections and point Removal Comparison }~(a,b): FedBary and Lava are superior; \textbf{Mislabeled samples detections and point Removal Comparison }~(c,d): FedBary performs similarly to Lava and conducts relatively accurate detections;} }
\label{combinedetect}
\vspace{-0.3cm} 
\end{figure}
\textcolor{black}{We plot noisy feature detections ($p=10$) with corresponding point removal experiments on CIFAR10 dataset in Figure~\ref{combinedetect}. The point removal experiment is performed with the following steps: removing data points from the entire training dataset in descending order of gradient values. Following the similar example in~\cite{jiang2023opendataval}, we use a logistic regression model with feature embeddings for this classification task. This process can be used to show the validation performance before and after removing training noisy points identified by FedBary. Specifically, once every more $5\%$ of datum is removed, we ft the model with the remaining dataset and evaluate its test accuracy on the holdout dataset. We visualize the accuracy w.r.t. fraction of removed valuable datasets on the right side~(b for noisy feature samples, d for mislabeled samples). }
\textcolor{black}{As for noisy feature detection~(a,b), our proposed approach is sensitive to noises in features based on the calibrated gradient, leading to superior performance in noisy features. The detection line of Lava is overlapping with ours thus we do not plot it~(note that in Lava raw data should be accessed). We also validate our statement in (e): Notably, the number of the positive gradient in $\partial \mathcal{W}_p(P_i,\eta_Q)$ is the same as the number of noisy data, representing our approach detects all noisy data without other clean data, which convincing its effectiveness. In (b) we could find Lava could detect valuable points that affect the model performance most while FedBary is runner-up with privacy guarantee.
For the label detection~(c,d), we observe the Wasserstein distances change trivially with different portions of label noise, thus making the detections relatively poor. However, FedBary performs similarly with approaches that could access data, indicating our approach does not sacrifice the performance of the benchmark. In (d), Lava performs slightly better, while KNNShapley gets stuck when about $55\%$ of the dataset is removed. LOO is relatively random in the experiment. Overall, we could say FedBary indeed provides valuable information as removing high negative values will lead to lower accuracy. \textcolor{black}{It is worthy to note that, for corrupt feature detections, FedBary performs 100$\%$ success rate~(only noisy data points and all of them have positive calibrated gradients) in various datasets, showing its generalization ability. }}\\
\textbf{Boost FL Model} We also simulate an FL setting, where $5000$ training samples are divided randomly and assigned to $5$ clients, in which there are $500$ noisy data in total. There are $1000$ validation data and $1000$ testing data held by the server. Our evaluation approach proceeds as follows: before training a federated model, the server calculates the Wasserstein distance using validation data with the local samples and filters noisy samples; the filtered new training data are used to collaboratively train a federated model. We compared various datasets, and the aggregated algorithm is FedAVG~\cite{mcmahan2017communication}. Our approach could help detect noisy data and boost FL model performance in Table~\ref{boostmodel} in Appendix~\ref{moreexpboost}.
\subsection{Duplication Robustness} 
One concern in real-world data marketplaces revolves around the ease of data duplication, which does not introduce any new information. \cite{xu2021validation,just2023lava} have emphasized the importance of a metric that can withstand data duplication. It is also likely that the client identifies the datum with the highest contribution and duplicates it in an attempt to maximize profit. FedBary is formulated in terms of distributions and automatically disregards duplicate sets. We conducted the experiment using CIFAR10, where we simulated $5000$ training data and $5000$ validation data. We repeated the training set up to three times, and the distance remained unchanged. We also duplicate a single datum with a large negative gradient value in~\ref{calivalue} multiple times for evaluation, while the result shows it would increase the distance due to the resulting imbalance in the training distribution caused by copying that particular point. We show the result in Table~\ref{robust}. We also discuss the time robustness in Appendix~\ref{timerobust}.
\begin{table}
    \centering
    \setlength{\tabcolsep}{0.8mm}
    \begin{tabular}{cc cc}
    \toprule[1pt] 
    &   & \multicolumn{2}{c}{Duplication}\\
     & Data Size  & Whole Dataset & One datum \\\hline
& 5000 & 39.14 &  40.94  \\
 &2 $\times$ 5000 & 39.23&	40.96   \\
& 3 $\times$ 5000 & 39.18&  40.99 \\
    \bottomrule[1pt]     
    \end{tabular}
    \caption{Distance behavior under duplications}
    \label{robust}
\vspace{-2.0em}
\end{table}
\section{Conclusion}
We propose Wasserstein distance in the FL context as a new metric for client evaluation and data detection. Compared to previous approaches, Our approach, \texttt{FedBary}, offers a more transparent and robust framework, substantiated through both theoretical and empirical analyses. In addition, it is designed to be applicable in real-world data marketplaces, enabling the data client evaluation before the FL training process. This facilitates the selection of only the most relevant clients and data points for the FL training process, thereby optimizing computational efficiency and enhancing model performance. We posit that such a metric holds considerable promise not only for the purpose of client evaluation but also as a cornerstone for developing incentive mechanisms within FL systems. While \texttt{FedBary} marks a significant advancement in this field, there remain several open questions and avenues for further research, which we have detailed in Appendix~\ref{openquestion}. 

\newpage
{
    \small
    \bibliographystyle{ieeenat_fullname}
    \bibliography{main}
}

\clearpage
\setcounter{page}{1}
\maketitlesupplementary

\section{Technical Concepts Definition}
\label{geodesic_def}

To formulate equation \ref{geodesicseq}, we utilized the mathematical property of Wasserstein Distance as below.
\begin{property}(Triangle Inequality of Wasserstein Distance)
\label{triangle inequality}
For any $p\geq 1, P,Q, \gamma \in \mathcal{P}_p(X)$, $\mathcal{W}_p$ is a metric on $\mathcal{P}_p(X)$, as such it satisfies the triangle inequality as~\cite{peyre2019computational}
\begin{equation}
     \mathcal{W}_p(P,Q) \leq \mathcal{W}_p(P,\gamma)+  \mathcal{W}_p(\gamma,Q),
\end{equation}
\end{property}
in order to attain equality, \textit{geodesics} and \textit{Interpolating point} are defined as structuring tools of metric spaces.
\begin{definition}(Geodesics~\cite{ambrosio2005gradient})
    Let $(\mathcal{X},d)$ be a metric space. A constant speed geodesic  $x:[0,1]\rightarrow \mathcal{X}$ between $x_0,x_1 \in \mathcal{X}$ is a continuous curve such that $\forall a,b\in[0,1], d\big(x(a),x(b)\big) = |a-b|\cdot d(x_0,x_1)$.
\end{definition}
\begin{definition}(Interpolating point~\cite{ambrosio2005gradient})
    Any point $x_t$ from a constant speed geodestic $(x(t))_{t\in[0,1]}$ is an interpolating point and verifies $d(x_0,x_1) = d(x_0,x_t)+d(x_t,x_1)$.
\end{definition}
The above definitions and properties are used to define the interpolating measure of the Wasserstein distance:
\begin{definition}(Wasserstein Geodesics, Interpolating measure~\cite{ambrosio2005gradient,kolouri2017optimal})
Let $P,Q \in \mathcal{P}_p(X)$ with $X\subseteq \mathbb{R}^d$ compact, convex and equipped with $\mathcal{W}_p$. Let $\pi^\star\in\Pi(P,Q)$ be an optimal transport plan between two distributions $P$ and $Q$. For $t\in [0,1]$, let $\gamma_t = (\pi_t)_{\#}\pi^\star$ where $\pi_t(x,y) = (1-t)x+ty$, i.e. $\gamma_t$ is the push-forward measure of $\pi^\star$ under the map $\pi_t$. Then, the curve $\bar{\mu}= (\gamma_t)_{t\in[0,1]}$ is a constant speed geodesic,  also called a Wasserstein geodesics between $P$ and $Q$.
\end{definition}

\section{Proof for Theorem~\ref{theorem1}}
\label{proof_theorem1}

Since the $Q^{(k)}$ is the wasserstein barycenter for all interpolating measures $\gamma_i^{(k)}$ where $i \in [1,N]$, we have
\begin{align}
 & A^{(k)} = \sum_i^N [\mathcal{W}_p(P_i, \gamma_i^{(k)}) + \mathcal{W}_p(Q^{(k)}, \gamma_i^{(k)})] \\
 & = \sum_i^N\mathcal{W}_p(P_i, \gamma_i^{(k)}) + \sum_i^N \mathcal{W}_p(Q^{(k)}, \gamma_i^{(k)}) \\
 & \leq \sum_i^N\mathcal{W}_p(P_i, \gamma_i^{(k)}) + \sum_i^N \mathcal{W}_p(Q^{(k-1)}, \gamma_i^{(k)})
\end{align}

Define the interpolating measure between $\gamma_i^{(k)}$ and $Q^{(k-1)}$ as $\eta_{Q_i}^{(k)^{'}}$, we have $\mathcal{W}_p(Q^{(k-1)}, \gamma_i^{(k)}) = \mathcal{W}_p(Q^{(k-1)}, \eta_{Q_i}^{(k)^{'}})+\mathcal{W}_p(\eta_{Q_i}^{(k)^{'}},\gamma_i^{(k)})$. \\
Based on Algorithm \ref{algorithm_1_fedbary}, $\eta_{P_i}^{(k+1)}$ is the interpolating measure for $P_i$ and $\gamma_i^{(k)}$. Thus, we can derive that,
\begin{align}
 & \mathcal{W}_p(P_i, \eta_{P_i}^{(k+1)}) + \mathcal{W}_p(\eta_{P_i}^{(k+1)},\gamma_i^{(k)}) \nonumber \\
 & \leq \mathcal{W}_p(P_i, \eta_{P_i}^{(k)}) + \mathcal{W}_p(\eta_{P_i}^{(k)},\gamma_i^{(k)}) \\
 & \mathcal{W}_p(Q^{(k-1)}, \eta_{Q_i}^{(k)^{'}})+\mathcal{W}_p(\eta_{Q_i}^{(k)^{'}},\gamma_i^{(k)}) \nonumber \\
 & \leq \mathcal{W}_p(Q^{(k-1)}, \eta_{Q_i}^{(k)}) + \mathcal{W}_p(\eta_{Q_i}^{(k)},\gamma_i^{(k)}) 
\end{align}
These two inequalities lead to
\begin{align}
 & \mathcal{W}_p(P_i, \eta_{P_i}^{(k+1)}) + \mathcal{W}_p(\eta_{P_i}^{(k+1)},\gamma_i^{(k)}) \nonumber \\
 & + \mathcal{W}_p(Q^{(k-1)}, \eta_{Q_i}^{(k)^{'}})+\mathcal{W}_p(\eta_{Q_i}^{(k)^{'}},\gamma_i^{(k)}) \nonumber \\
 & \leq \mathcal{W}_p(P_i, \eta_{P_i}^{(k)}) + \mathcal{W}_p(\eta_{P_i}^{(k)},\gamma_i^{(k)}) \\
 & + \mathcal{W}_p(Q^{(k-1)}, \eta_{Q_i}^{(k)}) + \mathcal{W}_p(\eta_{Q_i}^{(k)},\gamma_i^{(k)}) 
\end{align}
Simultaneously, the $\gamma_i^{(k)}$ is the interpolating measure for $\eta_{P_i}^{(k)}$ and $\eta_{Q_i}^{(k)}$. So we have
\begin{align}
 & \mathcal{W}_p(\eta_{P_i}^{(k)},\gamma_i^{(k)}) + \mathcal{W}_p(\eta_{Q_i}^{(k)},\gamma_i^{(k)}) \nonumber\\
 & \leq \mathcal{W}_p(\eta_{P_i}^{(k)},\gamma_i^{(k-1)}) + \mathcal{W}_p(\eta_{Q_i}^{(k)},\gamma_i^{(k-1)})
\end{align}
and
\begin{align}
 & \mathcal{W}_p(P_i, \eta_{P_i}^{(k+1)}) + \mathcal{W}_p(\eta_{P_i}^{(k+1)},\gamma_i^{(k)}) \nonumber \\
 & + \mathcal{W}_p(Q^{(k-1)}, \eta_{Q_i}^{(k)^{'}})+\mathcal{W}_p(\eta_{Q_i}^{(k)^{'}},\gamma_i^{(k)}) \nonumber \\
 & \leq \mathcal{W}_p(P_i, \eta_{P_i}^{(k)}) + \mathcal{W}_p(Q^{(k-1)}, \eta_{Q_i}^{(k)})  \nonumber\\
 & + \mathcal{W}_p(\eta_{P_i}^{(k)},\gamma_i^{(k-1)}) + \mathcal{W}_p(\eta_{Q_i}^{(k)},\gamma_i^{(k-1)}) \\
 & = \mathcal{W}_p(P_i, \gamma_i^{(k-1)}) + \mathcal{W}_p(Q^{(k-1)}, \gamma_i^{(k-1)})
\end{align}
Hence, we can now derive that
\begin{align}
 & A^{(k)} \leq \sum_i^N\mathcal{W}_p(P_i, \gamma_i^{(k)}) + \sum_i^N \mathcal{W}_p(Q^{(k-1)}, \gamma_i^{(k)}) \\
 &= \sum_i^N [\mathcal{W}_p(P_i, \eta_{P_i}^{(k+1)}) + \mathcal{W}_p(\eta_{P_i}^{(k+1)},\gamma_i^{(k)})] \nonumber\\
 & + \sum_i^N  [\mathcal{W}_p(Q^{(k-1)}, \eta_{Q_i}^{(k)^{'}})+\mathcal{W}_p(\eta_{Q_i}^{(k)^{'}},\gamma_i^{(k)})] \\
 & = \sum_i^N [\mathcal{W}_p(P_i, \eta_{P_i}^{(k+1)}) + \mathcal{W}_p(\eta_{P_i}^{(k+1)},\gamma_i^{(k)}) \nonumber\\
 & + \mathcal{W}_p(Q^{(k-1)}, \eta_{Q_i}^{(k)^{'}})+\mathcal{W}_p(\eta_{Q_i}^{(k)^{'}},\gamma_i^{(k)})]\\
 & \leq \sum_i^N [\mathcal{W}_p(P_i, \gamma_i^{(k-1)}) + \mathcal{W}_p(Q^{(k-1)}, \gamma_i^{(k-1)})] = A^{(k-1)} 
\end{align}
Thus, the sequence $A_{(k)}$ is non-increasing. By the triangle inequality, we have for any $k \in \mathbb{N}$,
\begin{align}
 \sum_i^N \mathcal{W}_p(P_i, Q) \leq A^{(k)}
\end{align}
Using the monotone convergence theorem, since $A^{(k)}$ is non-increasing and
bounded sequence below, then it converges to its infimum.

\section{Proof for Theorem~\ref{theorem2}}
\label{proof_theorem2}
In this section, we give a detailed proof for Theorem \ref{theorem2}, which is a restatement of the proof for Theorem 1 in \cite{just2023lava}. First, denote joint distribution of random data-label pairs $(x, f_t(x))_{x\sim P_i(x)}$ and $(x, f_v(x))_{x\sim Q(x)}$ as $P_i^{f_t}$ and $Q^{f_v}$ respectively, which are the same notation as $P_i$ and $Q$ but made with explicit dependence on $f_t$ and $f_v$ for clarity. The
distributions of $(f_t(x))_{x\sim P_i(x)}$ and $(f_v(x))_{x\sim Q(x)}$ as $P_{i_{f_t}}$ and $Q_{f_v}$ respectively. Besides, we define conditional distributions $P_i(x|y):=\frac{P_i(x)I[f_t(x)=y]}{\int P_i(x)I[f_t(x)=y]dx} $
and  $Q(x|y):=\frac{Q(x)I[f_v(x)=y]}{\int Q(x)I[f_v(x)=y]dx} $. Also, we denote $\pi \in \Pi(P_i,Q)$ as a coupling between a pair of distributions $P_i,Q$ and $d :\mathcal{X} \times \mathcal{X} \rightarrow \mathbb{R}$ as distance metric function. Generally, the $p$-Wasserstein distance with respect to cost function $\mathcal{C}$ is defined as $\mathcal{W}_p(P_i,Q):=\inf_{\pi\in\Pi(P_i,Q)}\mathbb{E}_{(x,y)\sim\pi}[\mathcal{C}(x,y)]$.

To prove Theorem \ref{theorem2}, the concept of probabilistic $cross$-Lipschitzness is needed, and it is assumed that
two labeling functions should produce consistent labels with high probability on two close instances.
\begin{definition}
\label{definition1}
(Probabilistic Cross-Lipschitzness). Two labeling functions $f_t:\mathcal{X}\rightarrow\{0,1\}^V$ and $f_v:\mathcal{X}\rightarrow\{0,1\}^V$ are $(\epsilon,\delta)$-probabilistic cross-Lipschitz w.r.t. a joint distribution $\pi$ over $\mathcal{X} \times \mathcal{X}$ if for all $\epsilon > 0$:
\begin{equation}
    P_{(x_1,x_2)\sim\pi}[||f_t(x_1)-f_v(x_2)|| > \epsilon d(x_1,x_2)] \leq \delta 
\end{equation}
\end{definition}
Given labeling functions $f_t$, $f_v$ and a coupling $\pi$, we can bound the probability of finding
pairs of training and validation instances labeled differently in a (1/$\epsilon$)-ball with respect to $\pi$.

Let $\pi_{x,y}^{\ast}$ be the coupling between $P_i^{f_t}$ and $Q^{f_v}$ such that 
\begin{align}
    &\pi_{x,y}^{\ast} := \nonumber\\
    & \arg_{\pi\in\Pi(P_i^{f_t},Q^{f_v})} \inf  \mathbb{E}_{((x_i,y_i),(x_q,y_q))\sim\pi}[\mathcal{C}((x_i,y_i),(x_q,y_q))].
\end{align}
We define two couplings $\pi^{\ast}$ and $\widetilde{\pi}^{\ast}$ between $P_i(x),Q(x)$ as follows:
\begin{equation}
    \pi^{\ast}(x_i,x_q) := \int_{\mathcal{Y}}\int_{\mathcal{Y}}\pi_{x,y}^{\ast}((x_i,y_i),(x_q,y_q))dy_idy_q.
\end{equation}
For $\widetilde{\pi}^{\ast}$, we first need to define a coupling between $P_{i_{f_t}}$ and $Q_{f_v}$:
\begin{equation}
    \pi_y^{\ast}(y_i,y_q) := \int_{\mathcal{X}}\int_{\mathcal{X}}\pi_{x,y}^{\ast}((x_i,y_i),(x_q,y_q))dx_idx_q
\end{equation}
and another coupling between $P_i^{f_t},Q^{f_v}$:
\begin{equation}
   \widetilde{\pi}_{x,y}^{\ast}(x_i,y_i),(x_q,y_q)) := \pi_y^{\ast}(y_i,y_q)P_i(x_i|y_i)Q(x_q|y_q).
\end{equation}
Finally, $\widetilde{\pi}^{\ast}$ is constructed as follows:
\begin{equation}
   \widetilde{\pi}^{\ast}(x_i,x_q) := \int_{\mathcal{Y}}\int_{\mathcal{Y}}\pi_y^{\ast}(y_i,y_q)P_i(x_i|y_i)Q(x_q|y_q)dy_idy_q.
\end{equation}

Next, we are going to prove Theorem \ref{theorem2}. The lefthand side of inequality in the theorem can be written as
\begin{align}
    & \mathbb{E}_{x\sim Q(x)}[\mathcal{L}(f_v(x),f(x))] \nonumber\\
    & = \mathbb{E}_{x\sim Q(x)}[\mathcal{L}(f_v(x),f(x))] \nonumber\\
    & -  \mathbb{E}_{x\sim P_i(x)}[\mathcal{L}(f_t(x),f(x))]  +  \mathbb{E}_{x\sim P_i(x)}[\mathcal{L}(f_t(x),f(x))] \\
    & \leq \mathbb{E}_{x\sim  P_i(x)}[\mathcal{L}(f_t(x),f(x))] \nonumber \\
    & + | \mathbb{E}_{x\sim Q(x)}[\mathcal{L}(f_v(x),f(x))] - \mathbb{E}_{x\sim P_i(x)}[\mathcal{L}(f_t(x),f(x))] |
\end{align}
We bound $| \mathbb{E}_{x\sim Q(x)}[\mathcal{L}(f_v(x),f(x))] - \mathbb{E}_{x\sim P_i(x)}[\mathcal{L}(f_t(x),f(x))] |$ as follows:
\begin{align}
    & | \mathbb{E}_{x\sim Q(x)}[\mathcal{L}(f_v(x),f(x))] - \mathbb{E}_{x\sim P_i(x)}[\mathcal{L}(f_t(x),f(x))] | \nonumber \\
    & = |\int_{\mathcal{X}^2}[\mathcal{L}(f_v(x_q),f(x_q))-\mathcal{L}(f_t(x_i),f(x_i))]d\pi^{\ast}(x_i,x_q)| \\
    & = |\int_{\mathcal{X}^2}[\mathcal{L}(f_v(x_q),f(x_q))-\mathcal{L}(f_v(x_q),f(x_i)) \nonumber \\
    & + \mathcal{L}(f_v(x_q),f(x_i)) -\mathcal{L}(f_t(x_i),f(x_i))]d\pi^{\ast}(x_i,x_q)| \\
    & \leq \underbrace{\int_{\mathcal{X}^2}|\mathcal{L}(f_v(x_q),f(x_q))-\mathcal{L}(f_v(x_q),f(x_i))|d\pi^{\ast}(x_i,x_q) }_{U_1} \nonumber \\
    & + \underbrace{\int_{\mathcal{X}^2}|\mathcal{L}(f_v(x_q),f(x_i)) - \mathcal{L}(f_t(x_i),f(x_i))|d\pi^{\ast}(x_i,x_q)}_{U_2}
\end{align}
where the last inequality is due to triangle inequality.
Now, we bound $U_1$ and $U_2$ separately. For $U_1$, we have
\begin{align}
    U_1 & \leq k\int_{\mathcal{X}^2}||f(x_q)-f(x_i)||d\pi^{\ast}(x_i,x_q) \\
    & \leq k\epsilon\int_{\mathcal{X}^2}\mathrm{d}(x_i,x_q)d\pi^{\ast}(x_i,x_q),
\end{align}
where both inequalities are due to Lipschitzness of $\mathcal{L}$ and $f$.
Recall that $\pi_y^{\ast}(y_i,y_q) := \int_{\mathcal{X}}\int_{\mathcal{X}}\pi_{x,y}^{\ast}((x_i,y_i),(x_q,y_q))dx_idx_q$ and $ \widetilde{\pi}_{x,y}^{\ast}(x_i,y_i),(x_q,y_q)) := \pi_y^{\ast}(y_i,y_q)P_i(x_i|y_i)Q(x_q|y_q)$. And for $U_2$, we can derive that
\begin{align}
    U_2 & \leq k \int_{\mathcal{Y}^2}\int_{\mathcal{X}^2}||y_q-y_i||d\pi_{x,y}^{\ast}((x_i,y_i),(x_q,y_q)) \\
    & = k \int_{\mathcal{Y}^2}||y_q-y_i||d\pi_y^{\ast}(y_i,y_q) \\
    & = k \int_{\mathcal{X}^2}\int_{\mathcal{Y}^2}||y_q-y_i||d\widetilde{\pi}_{x,y}^{\ast}((x_i,y_i),(x_q,y_q)) \\
    & = k \int_{\mathcal{Y}^2}\int_{\mathcal{X}^2}||f_v(x_q)-f_t(x_i)||d\widetilde{\pi}_{x,y}^{\ast}((x_i,y_i),(x_q,y_q)),
\end{align}
where the last step holds since if $y_i\neq f_t(x_i)$ or $y_q\neq f_v(x_q)$, then $\widetilde{\pi}_{x,y}^{\ast}((x_i,y_i),(x_q,y_q))=0$.
Define the region $\mathcal{A}={(x_i,y_i):||f_v(x_q)-f_t(x_i)||<\epsilon_{tv}\mathrm{d}(x_i,x_q)}$, then 
\begin{align}
    U_2 & \leq  k \int_{\mathcal{Y}^2}\int_{\mathcal{X}^2}||f_v(x_q)-f_t(x_i)||d\widetilde{\pi}_{x,y}^{\ast}((x_i,y_i),(x_q,y_q)) \\
    & = k \int_{\mathcal{Y}^2}\int_{\mathcal{X}^2\setminus \mathcal{A}}||f_v(x_q)-f_t(x_i)||d\widetilde{\pi}_{x,y}^{\ast}((x_i,y_i),(x_q,y_q)) \nonumber\\
    & + k \int_{\mathcal{Y}^2}\int_{\mathcal{A}}||f_v(x_q)-f_t(x_i)||d\widetilde{\pi}_{x,y}^{\ast}((x_i,y_i),(x_q,y_q)) \\
    & \leq k \int_{\mathcal{Y}^2}\int_{\mathcal{X}^2\setminus \mathcal{A}} 2V d\widetilde{\pi}_{x,y}^{\ast}((x_i,y_i),(x_q,y_q)) \nonumber\\
    & + k \int_{\mathcal{Y}^2}\int_{\mathcal{A}}||f_v(x_q)-f_t(x_i)||d\widetilde{\pi}_{x,y}^{\ast}((x_i,y_i),(x_q,y_q)).
\end{align}
Define $\widetilde{f}_t(x_i) = f_t(x_i) $ and $\widetilde{f}_v(x_q) = f_v(x_q) $ if $(x_i,x_q)\in \mathcal{A}$, and $\widetilde{f}_t(x_i) = \widetilde{f}_v(x_q) = 0 $ otherwise (note that $||\widetilde{f}_v(x_q)-\widetilde{f}_t(x_i)||<\epsilon_{tv}\mathrm{d}(x_i,x_q)$ for all $(x_i,x_q) \in \mathcal{X}^2$), then we can bound the second term as follows:
\begin{align}
    & k \int_{\mathcal{Y}^2}\int_{\mathcal{A}}||f_v(x_q)-f_t(x_i)||d\widetilde{\pi}_{x,y}^{\ast}((x_i,y_i),(x_q,y_q)) \\
    & \leq  k \int_{\mathcal{Y}^2}d\pi_y^{\ast}(y_i,y_q) \nonumber\\
    & \int_{\mathcal{A}}||f_v(x_q)-f_t(x_i)||dP_i(x_i|y_i)dQ(x_q|y_q) \\
    & = k \int_{\mathcal{Y}^2}d\pi_y^{\ast}(y_i,y_q) \nonumber \\
    & \int_{\mathcal{X}}^2||\widetilde{f}_v(x_q)-\widetilde{f}_t(x_i)||dP_i(x_i|y_i)dQ(x_q|y_q)\\
    & = k \int_{\mathcal{Y}^2}d\pi_y^{\ast}(y_i,y_q) \nonumber \\
    & \int_{\mathcal{X}}^2|| \mathbb{E}_{x_q\sim Q(\cdot|y_q)}[\widetilde{f}_v(x_q)]-\mathbb{E}_{x_i\sim P_i(\cdot|y_i)}[\widetilde{f}_t(x_i)]||\\
    & \leq k\epsilon_{tv}\int_{\mathcal{Y}^2}d\pi_y^{\ast}(y_i,y_q)\mathcal{W}_p(P_i(\cdot|y_i),Q(\cdot|y_q)).
\end{align}
The last inequality is a consequence of the duality form of the Kantorovich-rubinstein 
theorem \cite{villani2021topics}.
Combining all parts, we have
\begin{align}
   & U_1 + U_2  \leq k\epsilon\int_{\mathcal{X}^2}\mathrm{d}(x_i,x_q)d\pi^{\ast}(x_i,x_q)  \nonumber\\
   & +  k \int_{\mathcal{Y}^2}\int_{\mathcal{X}^2\setminus \mathcal{A}} 2V d\widetilde{\pi}_{x,y}^{\ast}((x_i,y_i),(x_q,y_q)) \nonumber\\
   & + k\epsilon_{tv}\int_{\mathcal{Y}^2}d\pi_y^{\ast}(y_i,y_q)\mathcal{W}_d(P_i(\cdot|y_i),Q(\cdot|y_q)) \\
   & \leq k\epsilon\int_{\mathcal{X}^2}\mathrm{d}(x_i,x_q)d\pi^{\ast}(x_i,x_q) + 2kV\delta_{tv} \nonumber\\
   &  + k\epsilon_{tv}\int_{\mathcal{Y}^2}d\pi_y^{\ast}(y_i,y_q)\mathcal{W}_p(P_i(\cdot|y_i),Q(\cdot|y_q)) \\
   & = 2kV\delta_{tv} + k\int_{(\mathcal{X} \times \mathcal{Y})^2}[\epsilon d(x_i,x_q)  \nonumber \\
   & + \epsilon_{tv}\mathcal{W}_p(P_i(\cdot|y_i),Q(\cdot|y_q))]d\pi_{x,y}^{\ast}((x_i,y_i),(x_q,y_q))  \\
   & \leq  2kV\delta_{tv} + k\int_{(\mathcal{X} \times \mathcal{Y})^2}[\epsilon d(x_i,x_q) + \nonumber \\
   & c\epsilon\mathcal{W}_p(P_i(\cdot|y_i),Q(\cdot|y_q))]d\pi_{x,y}^{\ast}((x_i,y_i),(x_q,y_q)) \\
   & = k\epsilon\mathbb{E}_{\pi_{x,y}^{\ast}}[\mathcal{C}((x_i,y_i),(x_q,y_q))] + 2kV\delta_{tv} \\
   & = k\epsilon\mathcal{W}_p(P_i^{f_t}, Q^{f_v}) + 2kV\delta_{tv}.
\end{align}
Thus the inequality in theorem \ref{theorem2} has been proved. For a more detailed discussion, please refer to \cite{just2023lava}.

\begin{figure*}
\setlength{\abovecaptionskip}{0 cm} 
\setlength{\belowcaptionskip}{0.1 cm} 
\centering
    \includegraphics[width = 1\textwidth]{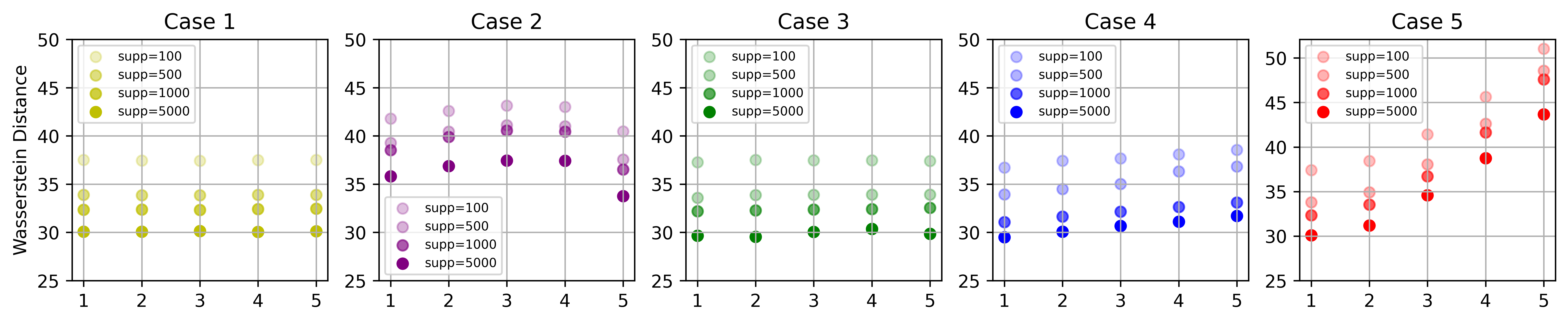}
    \caption{Wassestein Distances Under Different Support for $\textbf{Q}$ (x-axis is the client number)}
\label{wdfigure}
\end{figure*}

\section{Experiments Details}
\subsection{Client Evaluation}
We follow the same settings in~\cite{liu2022gtg} and~\cite{song2019profit} and divide the combination of distribution and size for client data into five different cases.
\label{clientvaldata}

\noindent\textbf{(1)~Same Distribution and Same Size:} All five clients possess the same number of images for each class;\\
\textbf{(2)~Different Distributions and Same Size:} Each participant has the same number of samples.
However, the Participant 1 dataset contains 80$\%$ for two classes. The other clients evenly
divide the remaining 20$\%$ of the samples. Similar procedures are applied to the rest;\\
\textbf{(3)~Same Distribution and Different Sizes:} Randomly sample from the entire training set
according to pre-defined ratios to form the local dataset for each participant, while ensuring
that there are the same number of images for each class in each participant. The
ratios for client 1-5 are: 10$\%$,15$\%$, 20$\%$, 25$\%$ and 30$\%$;\\
\textbf{(4)~Noisy Labels and Same Size:} Adopt the dataset from case (1), and flip the labels of a pre-defined percentage of samples in each participant’s local dataset. The
ratios for client 1-5 are: 0$\%$,5$\%$, 10$\%$, 15$\%$ and 20$\%$;\\
\textbf{(5)~Noisy Features and Same Size:} Adopt the dataset from case (1), and add different percentages of Gaussian noise into the input images. The
ratios for client 1-5 are: 0$\%$,5$\%$, 10$\%$, 15$\%$ and 20$\%$.

\subsection{Implementation Details}
The code implementation is developed based on Pytorch. We also developed our code and conduct comparisons based on the reference from the following sources:
\begin{itemize}
    \item \href{https://github.com/opendataval/opendataval/tree/main}{Opendataval~\cite{jiang2023opendataval}}
    \item \href{https://github.com/ruoxi-jia-group/LAVA}{LAVA~\cite{just2023lava}}
    \item \href{https://github.com/eddardd/WBTransport/tree/main}{WBTransport~\cite{montesuma2021wasserstein}}
    \item \href{https://github.com/cyyever/distributed_learning_simulator}{GTG-Shapley~\cite{liu2022gtg}}
\end{itemize}

\subsection{Experimental Baselines}

    \textbf{Original Shapley} The Shapley value is a concept from cooperative game theory that has been applied to machine learning to attribute a value to each feature (or client) in a predictive model. It provides a way to fairly distribute the ``contribution" of each player in a cooperative game. In the context of federated learning, it can be assumed that the prediction of a model in the server is the outcome of a game, and each client is a player contributing to that prediction. The Shapley value assigns a value to each client based on its marginal contribution to the model's prediction across all possible combinations of data for different clients. 
    
    Formally, a coalitional game is defined as: There is a set $N$ (of $n$ players) and a function $v$ that maps subsets of players to the real numbers: $2^{N} \rightarrow \mathbb{R}$, with  $v(\emptyset )=0$, where $\emptyset$ denotes the empty set. Then a general formula to calculate the Shapley value is provided as follows:
    \begin{equation}
        \varphi_i(v)=\sum_{S\subseteq N\setminus \{i\} }\frac{|S|!(n-|S|-1)!}{n!}(v (S\cup\{i\})-v(S) )
    \end{equation}

    where $n$ is the total number of players and the sum extends over all subsets $S$ of $N$ not containing player $i$.
    
    \noindent\textbf{MR/OR} ~\cite{song2019profit}  These two metrics are based on the contribution index, which is a concept proposed by the paper to replace the original Shapley value. Since direct computing of the contribution index can be time-consuming, two gradient-based methods are provided to reduce the time. The first one reconstructs models by updating the initial global model in federated learning with the gradients in different rounds. Then it calculates the contribution index by the performance of these reconstructed models. The second method calculates the contribution index in each round by updating the global model in the previous round with the gradients in the current round. Contribution indexes of multiple rounds are then added together with elaborated weights to get the final result.
    
    \noindent\textbf{TMC-Shapley} ~\cite{ghorbani2019data} The Truncated Monte Carlo Shapley (TMC-Shapley) is a data evaluation metric to quantify the value of each training datum to the predictor performance. The Monte Carlo and
    gradient-based methods are developed to efficiently estimate the value in practical settings. However, this metric only considers the context of supervised machine learning, and the privacy demand is neglected in the paper.
    
    \noindent\textbf{FedShapley} ~\cite{wang2020principled} This metric is proposed as a variant of the Shapley value amenable to federated learning. The key idea of the modification is to characterize the aggregate value of the set of clients in the same round through the model performance change caused by the addition of their data and then use the Shapley value to distribute the value of the set to each client. Compared with the canonical SV, it can be calculated without incurring extra communication costs and is capable of capturing the effect of participation order on data value.
    
    \noindent\textbf{GTG-Shapley} ~\cite{liu2022gtg} The Guided Truncation Gradient Shapley (GTG-Shapley) approach is a modification of the original Shapley value to address the challenge of significant computation costs in practice. It reconstructs federated learning models from gradient updates for Shapley value calculation instead of repeatedly training with different combinations of participants. A guided Monte Carlo sampling technique is introduced into the algorithm, enhancing the efficiency of calculation and reducing the computation costs.
    
\section{Discussions}
\label{discuss}
\subsection{Hyperparameter Analysis} Our observations have shown that epoch $K$ exerts minimal influence on the approximated distance. Conversely, with an increasing quantity of supports of the interpolating measure $S$, the approximated distance progressively approaches the exact distance. Consequently, in the context of evaluating relative contributions, a choice of few epochs and supports can effectively approximate the relative distance, leading to a reduction in computational complexity.
\subsection{Consistent evaluation time }
\label{timerobust}
 We find the truncation techniques in~\cite{ghorbani2019data,liu2022gtg} depend on the test performance, when the performance with a certain subset of clients is above the pre-specified threshold, contributions of remaining clients are assigned to 0 without additional evaluations. Therefore, the evaluation time varies with different truncation times and in the worst case the truncation will be conducted at the last round, making the complexity approaching $\mathcal{O}(2^N)$. In addition, the gradients in~\cite{song2019profit} with noisy data make MR and OR approaches have larger elapsed time than other cases. However, FedBary is robust and the elapsed time will not be affected by data characteristics. 

\subsection{Client detection is better than server detection}
\label{more_exp}
The detection accuracy is $100\%$ in the client side with $\nabla \mathcal{W}_p(P_i,\eta_Q)$ and $76\%$ in the server side with $\nabla \mathcal{W}_p(\eta_Q,Q)$. We conjecture this result is due to the gradient towards the  $P_i$ is more informative and straightforward.  For the mislabeled data detection, our approach could only detect $45\%$ of noisy data. However, it is worth noting that when accessing data, the mislabeled detection accuracy is only $47\%$, and the bottom plots show two approaches are almost overlapping, which shows Fedbary does not sacrifice much performance on detections on the benchmark of accessing data.

\subsection{Boost FL Performance}
\label{moreexpboost}
\begin{table}
    \centering
    \setlength{\tabcolsep}{0.8mm}
    \begin{tabular}{ccccccc}
    \toprule[1pt] 
  Data &$\#$Noisy & $\#$Removed  & acc.before &  acc.after \\\hline
 CIFAR10  &  500 &  494 & 0.67 &  0.73  \\
Fashion & 1000  & 580   & 0.56  & 0.64 \\
    \bottomrule[1pt]     
    \end{tabular}
    \caption{Accuracy before$\slash$after removing detected noisy samples}
    \label{boostmodel}
    \vspace{-0.5em}
\end{table}
\subsection{Future Explorations}
\label{openquestion}
\textbf{Noisy Label Detection} The current approach, which relies on an augmented matrix based on a Gaussian approximation for the conditional distribution, demonstrates relatively poor performance in differentiating clean subsets from mislabeled ones compared to the case of noisy features. To address this issue, a potential future direction is to implement exact calculations for the Wasserstein distance by utilizing an interpolating measure and an appropriate embedding approach, to design a filtering approach for the mislabeled case.  \\
\textbf{Accurate Server Detection} 
Detecting noisy data from the server side is crucial for defending against potential attacks from untrusted clients. The current approach, primarily driven by client-side analysis, excels in detection due to the client's access to their own data and the ability to measure gradients with respect to the interpolating measure such as $\gamma_i$ or  $\eta_{Q_i}$ shared from the server side. However, the detection ability from the server side is limited since it cannot access local client data, and using $\eta_{P_i}$ from the local client or $\gamma_i$ is less effective in our explorations. Strengthening the server-side detection capabilities is of paramount importance in the context of the security application. 

\end{document}